\documentclass{article}
\usepackage{mathtools} 
\usepackage{times}
\usepackage{graphicx} 
\usepackage{booktabs}
\usepackage{flexisym}
\usepackage{subcaption}

\usepackage{natbib}

\usepackage{algorithm}
\usepackage{algorithmic}
\usepackage{caption}

\usepackage{hyperref}

\usepackage[accepted]{icml2017}

\begin{document} 

\twocolumn[
\icmltitle{Multi-task Learning for Continuous Control}



\icmlsetsymbol{equal}{*}

\begin{icmlauthorlist}
\icmlauthor{Himani Arora}{equal,cu}
\icmlauthor{Rajath Kumar}{equal,cu}
\icmlauthor{Jason Krone}{equal,cu}
\icmlauthor{Chong Li}{cu}
\end{icmlauthorlist}

\icmlaffiliation{cu}{Columbia University, USA}

\icmlcorrespondingauthor{Jason Krone}{jpk2151@columbia.edu}
\icmlcorrespondingauthor{Himani Arora}{ha2434@columbia.edu}
\icmlcorrespondingauthor{Rajath Kumar}{rm3497@columbia.edu}

\icmlkeywords{boring formatting information, machine learning, ICML}
\vskip 0.3in
]


\printAffiliationsAndNotice{\icmlEqualContribution} 
\begin{abstract}
Reliable and effective multi-task learning is a prerequisite for the development of robotic agents that can quickly learn to accomplish related, everyday tasks.
However, in the reinforcement learning domain, multi-task learning has not exhibited the same level of success
as in other domains, such as computer vision. In addition, most reinforcement learning research on multi-task learning
has been focused on discrete action spaces, which are not used for robotic control in the real-world. 
In this work, we apply multi-task learning methods to continuous action spaces and 
benchmark their performance on a series of simulated continuous control tasks. Most notably, we show that multi-task
learning outperforms our baselines and alternative knowledge sharing methods. 
\end{abstract} 

\section{Introduction}
Currently, reinforcement learning algorithms are sample inefficient and
learn from scratch through trail and error over millions of rollouts.
This sample inefficiency is not a problem when the goal is to maximize performance on a single task in a
 simulated environment, where data is cheap and can be collected quickly. However, this inefficiency is not viable
in real-world use cases when the goal is for an agent to accomplish many tasks, which may change over time. In addition to developing more efficient algorithms, 
one solution to this problem is to share knowledge between multiple tasks and develop flexible representations that can easily transfer to new tasks. 
In this work, we take a prerequisite step in this direction by evaluating the performance of the state-of-the-art multi-task learning methods on continuous action spaces using an extended version of the MuJoCo environment \cite{four}. Analyzing the success of various multi-task methods on continuous control tasks is an important contribution because most research in this area has been on discrete action spaces in Atari environments.

\section{Related Work}
In recent years, a number of works, most consistently from Deep Mind, have proposed methods for
transfer learning and muli-task learning. 
These works primarily focus on two approaches: knowledge distillation and feature reuse. 
Knowledge distillation, originally proposed in \cite{seven}, serves as the foundation
for the methods put forward in \cite{two}, \cite{one}, and \cite{five}.
\cite{two} extends the distillation method, formulated in \cite{six},
to Deep Q Networks trained on Atari environments and demonstrates that policy distillation
can act as a form of regularization for Deep Q Networks. In \cite{one}, the authors propose a novel
loss function that includes both a policy regression term as well as a feature regression
term. This policy regression objective is traditionally used in distillation, while
the added regression objective encourages feature representations in the intermediate layers of the student network
to match those of the expert network.
\cite{five} applies distillation to the multi-task setting by learning a common (distilled)
policy across a number of 3D environments. In addition, \cite{five} adds an entropy penalty as well as
 $KL$ and entropy regularization coefficients to the objective in order to trade off between exploration and exploitation.
\cite{three} attack the problem of catastrophic forgetting, where a policy losses the ability to
preform a pre-transfer task after being transferred to a target task. Concretely, the authors prevent
catastrophic forgetting by maintaining task specific representations within the policy network.
The environments we use in our experiments were introduced in \cite{four}. 
These environments extend the Mujoco continuous control tasks available in Open AI Gym and are designed
to be a test bed for transfer learning and multi-task learning. For a given simulated agent, these environments provide minor structural variations such as the length of an agent's body parts.
\begin{figure*}
\centering
\begin{subfigure}[t]{0.45\linewidth}
\centering
\includegraphics[width=\linewidth]{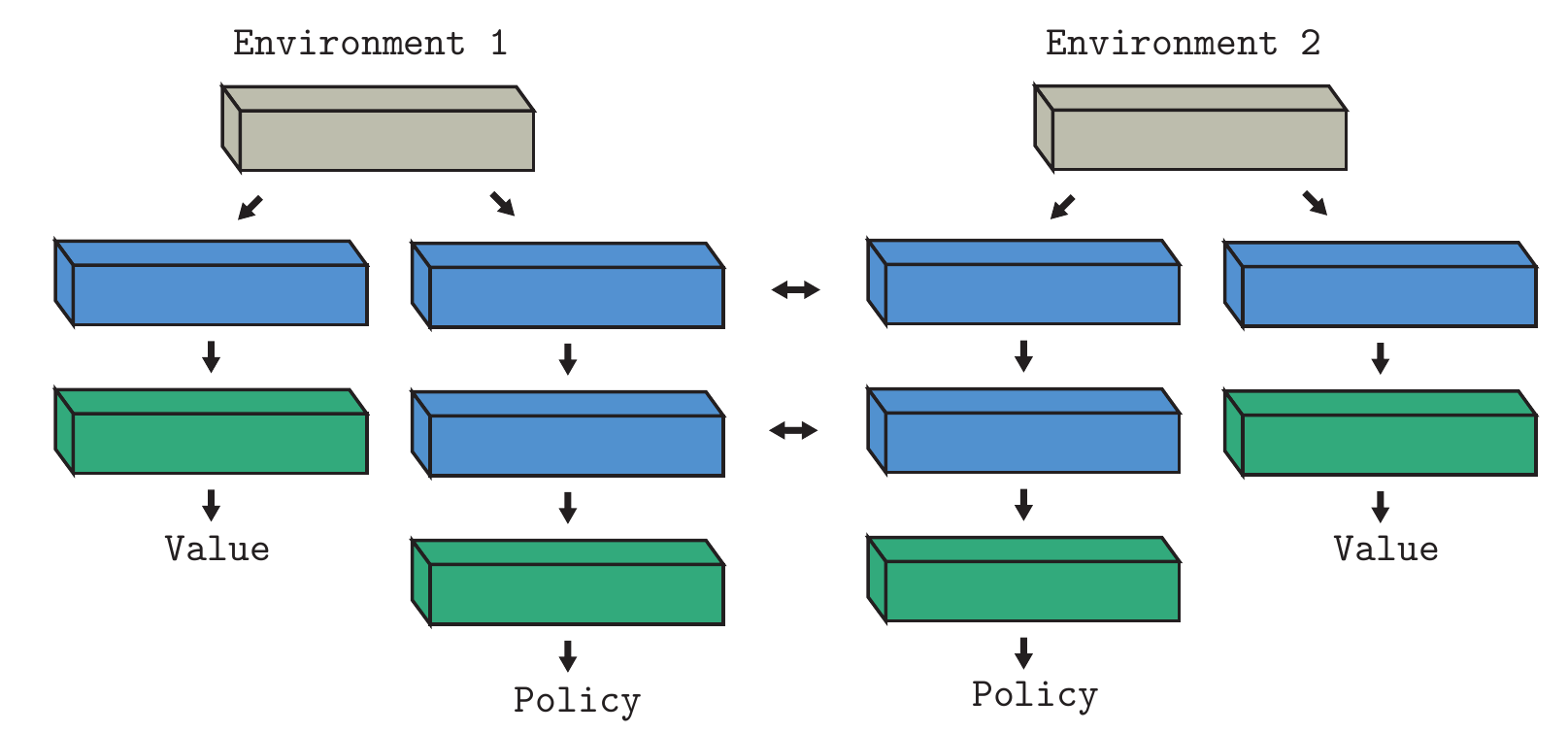}
\caption{Vanilla Multi-task}
\end{subfigure}
\hfill
\begin{subfigure}[t]{0.45\linewidth}
\centering
\includegraphics[width=\linewidth]{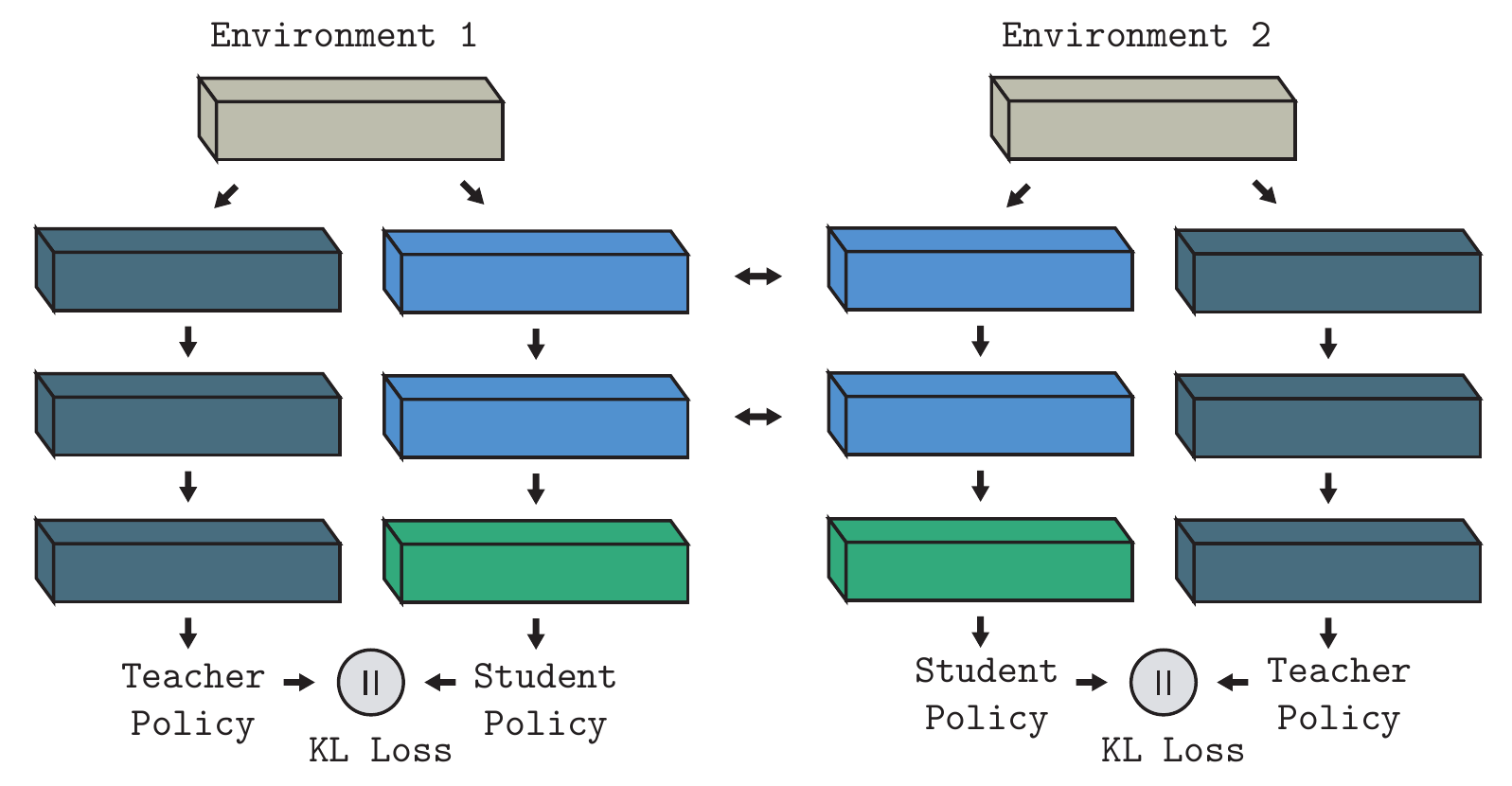}
\caption{Distillation Multi-Task}
\end{subfigure}
\caption{Proposed approach for multi-task learning. (a) Vanilla Multi-task (b) Distillation Multi-task. Green-- Trainable weights; Dark Grey-- Fixed Weights; Blue-- Shared Weights; Dark Blue-- Expert network Weights}
\label{methods_fig}
\end{figure*}

\section{Approach}
\subsection{Preliminaries}
\subsubsection{Markov Decision Process}
To provide context we give a brief review of the reinforcement learning problem. Reinforcement learning is the process of training
an agent to maximize reward in an environment. More technically, the aim is to learn the optimal
policy for selecting actions to take in a Markov Decision Process (MDP). A MDP is defined as
$\left< S, A, P, r, \rho_0, \gamma, T \right>$ where $S$ is a set of states, $A$
is a set of actions, $P : S \times A \times S \rightarrow {R_+}$ is a transition
function, $r : S \times A \rightarrow \left[ R_{min}, R_{max} \right]$ is a reward function,
$\gamma \in \left[0, 1 \right]$ is a discount factor, and $T$ is a time horizon. The policy
$\pi : S \times A \rightarrow {R_+}$ is trained to maximize the expected discounted
return $\eta(\pi_\theta) = {E}_{\tau \sim \pi_\theta(\tau)}\left[\sum_{t = 0}^{T} \gamma^tr(s_t, a_t) \right]$, where $\tau$
denotes a trajectory $\tau = (s_0, a_0, \dots)$ sampled according to $\pi_\theta$ with
$s_0 \sim \rho_0(s_0), a_t \sim \pi_\theta(a_t \mid s_t)$, and $s_{t+1} \sim P(s_{t+1} \mid s_t, a_t)$.
Furthermore, we define the optimal approximate policy as $\pi_{\theta^*}(a \vert s)$ where, 
\begin{equation}
\theta^* = \arg\max_{\theta} E_{\tau \sim \pi_\theta(\tau)}\left[\sum_{t = 0}^{T} \gamma^tr(s_t, a_t) \right]
\end{equation}
\begin{equation}
\pi_\theta(\tau) = p(s_1) \prod_t^T \pi_\theta(a_t \vert s_t)T(s_{t+1} \vert s_t, a_t)    
\end{equation}

\subsubsection{Actor Critic Algorithm}
To learn this optimal approximate policy we use the advantage actor-critic (A2C) algorithm, which is a synchronous implementation of the A3C algorithm introduced in \cite{eight}. A2C is an on-policy algorithm that operates in the forward view by sampling rollouts of current policy to calculate n-step returns. The policy as well as the value function are
updated after every $\mathrm{t_{max}}$ forward steps or when a terminal state is reached.
This algorithm maintains a policy $\pi_\theta$ and a value function estimate $V_{\theta_V}$ and performs updates of the form $\nabla_\theta \pi_\theta(a_t \vert s_t)A_{\theta, \theta_V}(s_t, a_t)$, where the advantage $A_{\theta, \theta_V}(s_t, a_t)$ is defined as
\begin{equation}
A_{\theta, \theta_V}(s_t, a_t) = \sum_{i=0}^{k-1} \gamma^ir_{t+i} + \gamma^kV_{\theta_V}(s_{t+k}) - V_{\theta_V}(s_t)
\end{equation}
We discourage convergence to sub-optimal deterministic policies can by adding a policy entropy term to the objective function as originally proposed in \cite{ten}. To approximate $\pi_\theta$ and $V_{\theta_V}$ we use feed forward neural networks. Since our experiments deal with continuous environments, we select actions by sampling from a multivariate Gaussian distribution with mean $\mu$ and diagonal co-variance matrix $\Sigma$ . Thus, the output layer of our policy network consists of a real-valued mean and the log variance for each dimension of the action space. 

\subsubsection{Knowledge Distillation}
\label{distil}
The goal of knowledge distillation is to transfer knowledge from a teacher model $T$ to a student model $S$. In our experiments, $T$ is a policy $\pi_\theta$ trained from scratch on a single environment using A2C, and $S$ is a feed forward network which has not been trained. $S$ is trained on the dataset $D = (X, Y)$, where $X$ denotes features and $Y$ denotes targets. $X$ contains state action pairs $(s_i, a_i)$ taken from trajectories $\tau = (s_0, a_0, \dots)$ of length $\mathrm{t_{max}}$, which are sampled according to the student's policy $\pi_{\theta_S}$ with probability $p_s$ and according to the teacher's policy $\pi_{\theta_T}$ with probability $p_t$. $Y$ contains the values $\mu_T, \Sigma_T = \pi_{\theta_T}(s_i, a_i)$ that parameterize the teacher's policy for the given state, action pair. We train $S$ on $D$ using the KL Divergence $\mathrm{D}_{\mathrm{KL}}$ between the teacher policy and the student policy as the objective function. Specifically, we use $L = \mathrm{D}_{\mathrm{KL}}(\mathcal{N}(\mu_{T}, \Sigma_{T}), \mathcal{N}(\mu_{S}, \Sigma_{S}))$ since the actions taken by $T$ and $S$ are drawn from multi-variate Gaussians. $\mathrm{D}_{\mathrm{KL}}$ is defined as follows:
\begin{equation}
\begin{multlined}
\mathrm{D}_{\mathrm{KL}}(\mathcal{N}(\mu_{T}, \Sigma_{T}), \mathcal{N}(\mu_{S}, \Sigma_{S})) = \\
\frac{1}{2}\left[
\log{\frac{|\Sigma_T|}{|\Sigma_S|}} - p + \mathrm{tr}(\Sigma_T^{-1}\Sigma_S)\right] + \\
\frac{1}{2}({\mu_T - \mu_S})^T\Sigma_T^{-1}({\mu_T - \mu_S})
\end{multlined}
\end{equation}
We choose to use KL Divergence as our loss function because it was shown to perform well on discrete action spaces in \cite{two}

\subsubsection{Multi-Task Learning}
To goal of multi-task learning  is to train a policy network $\pi_\theta$ that behaves optimally in $n$ different environments $E_1, E_2 \cdots E_n$. In our multi-task experiments, we approximate the optimal policy using an actor network that is essentially a feed forward network that contains two hidden layers shared across all environments and $n$ output layers (heads), where head $h_i$ produces the mean $\mu_i$ and covariance $\Sigma_i$ that parameterize the Gaussian policy for environment $E_i$. We experiment with two methods for training $\pi_\theta$ on multiple tasks: 
\begin{itemize}
    \item Vanilla multi-task learning: Each head $h_i$ is trained using A2C to maximize the expected discounted return for environment $E_i$. The value network in this case consists of one shared hidden layer and $n$ output head. During training, we sample an equal number of rollouts by cycling between  environment $E_1$ to environment $E_n$. We provide an illustration of vanilla multi-task in part (a) of figure 1. 
    
    \item Muti-task distillation: Each head $h_i$ is trained using knowledge distillation to match the output of a teacher network $T_i$. The Multi-task distillation training process is identical to the knowledge distilation process except that a dataset $D_i$ is collected for each teacher, head pair $(T_i, h_i)$ where the rollouts are sampled from the student network. We provide an illustration of muti-task distillation in part (b) of figure 1.
\end{itemize}
The above 2 methods for multi-task learning are illustrated in figure \ref{methods_fig}. 

\section{Experiments}
We conducted our experiments using the half-cheetah agent on 6 morphologically modified variants of the Open AI gym extensions described in \cite{four} namely \textit{HalfCheetahSmallFoot-v0}, \textit{HalfCheetahSmallLeg-v0}, \textit{HalfCheetahSmallTorso-v0} and \textit{HalfCheetahSmallThigh-v0} which reduce the size of the agent's respective body part by $25\%$ as well as on \textit{HalfCheetahBigFoot-v0} and \textit{HalfCheetahBigTorso-v0} which increase the size of the agent's respective body part by $25\%$. We evaluate the performance of a trained policy by reporting the mean and standard deviation of the cumulative reward across 20 sample rollouts on each target environment as done in \cite{four}. In addition, we plot the learning curves for each method in order to determine the sample efficiency of these approaches, which we provide in the appendix.

\subsection{Implementation}
We use PyTorch to implement all our models. Our actor network and critic networks consist of 2 and 3 fully-connected layers respectively, each of which have 64 hidden units. Because the Mujoco environments we use are for continuous control each action taken by an agent is sampled from a Gaussian distribution parameterized by the mean and variance given by $\pi_\theta$. We use RMSprop with an initial learning rate of 0.0007 to train our models. We set the A2C hyper parameter $\mathrm{t_{max}} = 5$ for all of our experiments. In addition, we use an entropy penalty coefficient of 0.01.  

\subsection{Fine-tuning}

\begin{table}[h!]
\centering
\resizebox{0.48\textwidth}{!}
{
\begin{tabular}{cccc}
\toprule
Training & Steps & $E_1$ & $E_2$ \\
\midrule
Scratch & 5M & 1582.78$\pm$30.46 & 1536.16$\pm$30.26 \\ 
\midrule
$E_1 \rightarrow E_2$ & 1M & 1530.26$\pm$29.22 & 1532.70 $\pm$ 23.35 \\
$E_1 \rightarrow E_2$ & 5M & 1500.76$\pm$30.36 & \textbf{1568.66 $\pm$ 29.56}\\
\midrule
$E_2 \rightarrow E_1$  & 1M & \textbf{1600.31 $\pm$ 26.32} & 1498.45$\pm$19.76\\
$E_2 \rightarrow E_1$  & 5M & 1493.02 $\pm$ 27.21 & 1425.64$\pm$29.82\\
\bottomrule
\end{tabular}}
\caption{$E_1$ and $E_2$ refer to \textit{HalfCheetahSmallFoot-v0} and \textit{HalfCheetahSmallLeg-v0} respectively. $E_1 \rightarrow E_2$ denotes finetuning the weights of $E_1$ on $E_2$ and vice-versa for $E_2 \rightarrow E_1$. Average and standard deviation ($\mu \pm \sigma$) of reward is calculated across a set of 20 sample rollouts. Steps is denoted in millions.}
\label{finetune_table}
\end{table}

\begin{table*}[h!]
\centering
\resizebox{0.8\textwidth}{!}{
\begin{tabular}{cccc}
\toprule
Environment & Scratch (3M) & Distillation Multi-task (1M) & Vanilla Multi-task (1M) \\
\midrule
HalfCheetahSmallFoot-v0 & 1326.30$\pm$21.15 & 1348.30$\pm$16.65 & \textbf{1480.89$\pm$30.92}\\
HalfCheetahBigFoot-v0 & 37.80$\pm$441.49 & -63.07$\pm$0.93 & \textbf{2188.11$\pm$21.28}\\
HalfCheetahSmallLeg-v0 & 1420.77$\pm$21.66 & 1413.55$\pm$17.73 & \textbf{1525.09$\pm$25.15}\\
HalfCheetahSmallThigh-v0 & 1444.21$\pm$24.75 & 838.35$\pm$960.78 & \textbf{1874.04$\pm$14.02} \\
HalfCheetahSmallTorso-v0 & 2116.10$\pm$29.24 & 2116.94$\pm$38.00 & \textbf{2276.11$\pm$46.14}\\
HalfCheetahBigTorso-v0 & 1312.50$\pm$24.75 & 1323.61$\pm$29.64 & \textbf{1428.97$\pm$7.06}\\
\bottomrule
\end{tabular}}
\caption{Results on multi-task network on 6 environments}
\label{multi_task_table}
\end{table*}

One of the simplest methods for multi-task learning is fine-tuning. For this, we first trained a policy with random initialization of weights for 5M frames on each environment separately. We then transferred this policy $\pi_\theta$ to another environment by initializing its weights of a new network $\pi_\theta'$ to the weights used in $\pi_\theta$ and then fine-tuning the last layer of the actor and critic networks in $\pi_\theta'$ for another 5M frames. We conducted our fine-tuning experiments on \textit{HalfCheetahSmallFoot-v0} and \textit{HalfCheetahSmallLeg-v0} and evaluated the results both at the 1M and 5M mark. The learning curves for all tasks are shown in Figure \ref{finetune_fig}.The mean and standard deviation of the accumulated rewards calculated on 20 rollouts of the policy are tabulated in Table \ref{finetune_table} for each combination of original and target environment. 

From the table it is clear that although fine-tuning the weights of a pre-trained network on a new environment performs better than training from scratch, its performance degrades on the original environment. Thus it suffers from catastrophic forgetting, which  makes it a poor choice for multi-task learning. For this reason, we did not
explore all combinations of original and target environments and instead focus on other types of multi-task learning, which we discuss below. 

\subsection{Multi-task learning}
\subsubsection{Vanilla multi-task learning}
In the vanilla multi-task experiment, we train a six head actor network and a six head critic network on each environment. Initial hidden layers are shared across environments, while output layers are unique to each environment's head as shown in Fig. \ref{methods_fig}. The training procedure is as follows, first sample rollouts from $E_1$  are collected and the head corresponding to the environment $E_1$ is trained, similarly $E_2$ and so on. the environments are continuously cycled in this manner until each head is trained for 1M frames. 
\\All results are tabulated in \ref{multi_task_table}. As clearly visible, the vanilla multi-task outperforms not only distillation multi-task but also networks trained from scratch on a single environment. This shows that sharing knowledge across multiple tasks helps the network perform better on each individual task as well re-affirming our original motivation for this work. In addition, it also helps train the network faster and achieves comparable performance in just 1M frames as compared to the 3M frames. 

\subsubsection{Multi-task distillation}
For the multi-task distillation experiment we first trained teacher networks on all 6 tasks separately for 3M frames each. Our multi-task network then consisted only of an actor network with shared hidden layers and 6 output head layers unique to each environment Fig. \ref{methods_fig}. The training procedure was similar to the vanilla multi-task except we sample rollouts from the student policy and used the knowledge distillation loss for training our network.
\\The motivation for exploring the use of distillation to train each head is two fold. Firstly, we hoped that distillation would
decrease training time by providing more stable targets for the actor and critic networks. Secondly, we thought that distillation
had the potential to stabilize the training process by mimicking the behavior of the expert teacher network. In the multi-task
learning graphs provided in section B of our appendix, you can see that our first assertion was correct. Namely, the reward of the
multi-task distillation agent reaches 1000 more quickly than the Vanilla distillation agent. However, the variance in the reward of 
the trained distillation agent is not notably lower than the variance of the vanilla distillation agent. This is evidenced by the fact
that the standard deviation of the reward of the distillation agent on $HalfCheetahBigTorso$ and $HalfCheetahSmallThigh$ is considerably 
larger than the standard deviation of the vanilla agent as shown in table \ref{multi_task_table}. 

\section{Conclusion}
In this paper, we experiment with two different methods for multi-task learning for continuous control. We show that an agent that is trained simultaneously to perform on multiple tasks is not only able to generalize better on each individual task but also requires fewer training steps to achieve comparable performance. This has a huge advantage since most of the real-world environments are continuous and sampling large numbers of episodes from them can be difficult. More sophisticated techniques could be developed that use a reinforcement learning policy to select environments to sample episodes from, which are likely to help the network focus on more difficult tasks and train faster. We hope our methods can serve as benchmark for future work in this field.

\section{Reproducibility}
We have provided all the trained model weights and codes at \textit{https://github.com/jasonkrone/elen6885-final-project}

\section{Acknowledgments}
The authors would like to thank Prof. Chong Li for sharing his knowledge and the Teaching Assistants-- Lingyu Zhang, Chen-Yu Yen and Xing Yuan for their constant support through out the course.

\bibliography{example_paper}

\begin{thebibliography}{9}
\providecommand{\natexlab}[1]{#1}
\providecommand{\url}[1]{\texttt{#1}}
\expandafter\ifx\csname urlstyle\endcsname\relax
  \providecommand{\doi}[1]{doi: #1}\else
  \providecommand{\doi}{doi: \begingroup \urlstyle{rm}\Url}\fi

\bibitem[Bucila et~al.(2006)Bucila, Caruana, and Niculescu-Mizil]{seven}
Bucila, Cristian, Caruana, Rich, and Niculescu-Mizil, Alexandru.
\newblock Model compression.
\newblock \emph{ACM}, 2006.

\bibitem[Henderson et~al.(2017)Henderson, Chang, Shkurti, Hansen, Meger, and
  Dudek]{four}
Henderson, Peter, Chang, Wei-Di, Shkurti, Florian, Hansen, Johanna, Meger,
  David, and Dudek, Gregory.
\newblock Benchmark environments for multitask learning in continuous domains.
\newblock \emph{arXiv preprint arXiv:1708.04352v1}, 2017.

\bibitem[Hinton et~al.(2014)Hinton, Vinyals, and Dean]{six}
Hinton, G., Vinyals, O., and Dean, J.
\newblock Distilling the knowledge in a neural network.
\newblock \emph{Deep Learning and Representation Learning Workshop, NIPS},
  2014.

\bibitem[Mnih et~al.(2016)Mnih, Badia, Mirza, Graves, Harley, Lillicrap,
  Silver, and Kavukcuoglu]{eight}
Mnih, Volodymyr, Badia, Adrià~Puigdomènech, Mirza, Mehdi, Graves, Alex,
  Harley, Tim, Lillicrap, Timothy~P., Silver, David, and Kavukcuoglu, Koray.
\newblock Asynchronous methods for deep reinforcement learning.
\newblock \emph{JMLR}, 2016.

\bibitem[Parisotto et~al.(2016)Parisotto, Ba, and Salakhutdinov]{one}
Parisotto, Emilio, Ba, Jimmy, and Salakhutdinov, Ruslan.
\newblock Actor-mimic deep multitask and transfer reinforcement learning.
\newblock \emph{ICLR 2016}, 2016.

\bibitem[Rusu et~al.(2016{\natexlab{a}})Rusu, Colmenarejo, Gulcehr, Desjardins,
  Kirkpatrick, Pascanu, Mnih, Kavukcuoglu, and Hadsell]{two}
Rusu, Andrei~A., Colmenarejo, Sergio~Gomez, Gulcehr, Caglar, Desjardins,
  Guillaume, Kirkpatrick, James, Pascanu, Razvan, Mnih, Volodymyr, Kavukcuoglu,
  Koray, and Hadsell, Raia.
\newblock Policy distilation.
\newblock \emph{arXiv preprint arXiv:1511.06295v2}, 2016{\natexlab{a}}.

\bibitem[Rusu et~al.(2016{\natexlab{b}})Rusu, Rabinowitz, Desjardins, Soyer,
  Kirkpatrick, Kavukcuoglu, Pascanu, and Hadsell]{three}
Rusu, Andrei~A., Rabinowitz, Neil~C., Desjardins, Guillaume, Soyer, Hubert,
  Kirkpatrick, James, Kavukcuoglu, Koray, Pascanu, Razvan, and Hadsell, Raia.
\newblock Progressive neural networks.
\newblock \emph{arXiv preprint arXiv:1606.04671v3}, 2016{\natexlab{b}}.

\bibitem[Teh et~al.(2017)Teh, Bapst, Czarnecki, Quan, Kirkpatrick, Hadsell,
  Heess, and Pascanu]{five}
Teh, Yee~Whye, Bapst, Victor, Czarnecki, Wojciech~Marian, Quan, John,
  Kirkpatrick, James, Hadsell, Raia, Heess, Nicolas, and Pascanu, Razvan.
\newblock Distral: Robust multitask reinforcement learning.
\newblock \emph{arXiv preprint arXiv:1707.04175v1}, 2017.

\bibitem[Williams \& Peng(1991)Williams and Peng]{ten}
Williams, Ronald~J and Peng, Jing.
\newblock Function optimization using connectionist reinforcement learning
  algorithms.
\newblock \emph{Connection Science}, 3\penalty0 (3):\penalty0 241--268, 1991.

\end{thebibliography}
\bibliographystyle{icml2017}

\onecolumn
\appendix
\section{Fine-tuning Learning curves}

\begin{figure}[H]
\centering
\begin{subfigure}[t]{0.48\linewidth}
\centering
\includegraphics[width=0.8\linewidth]{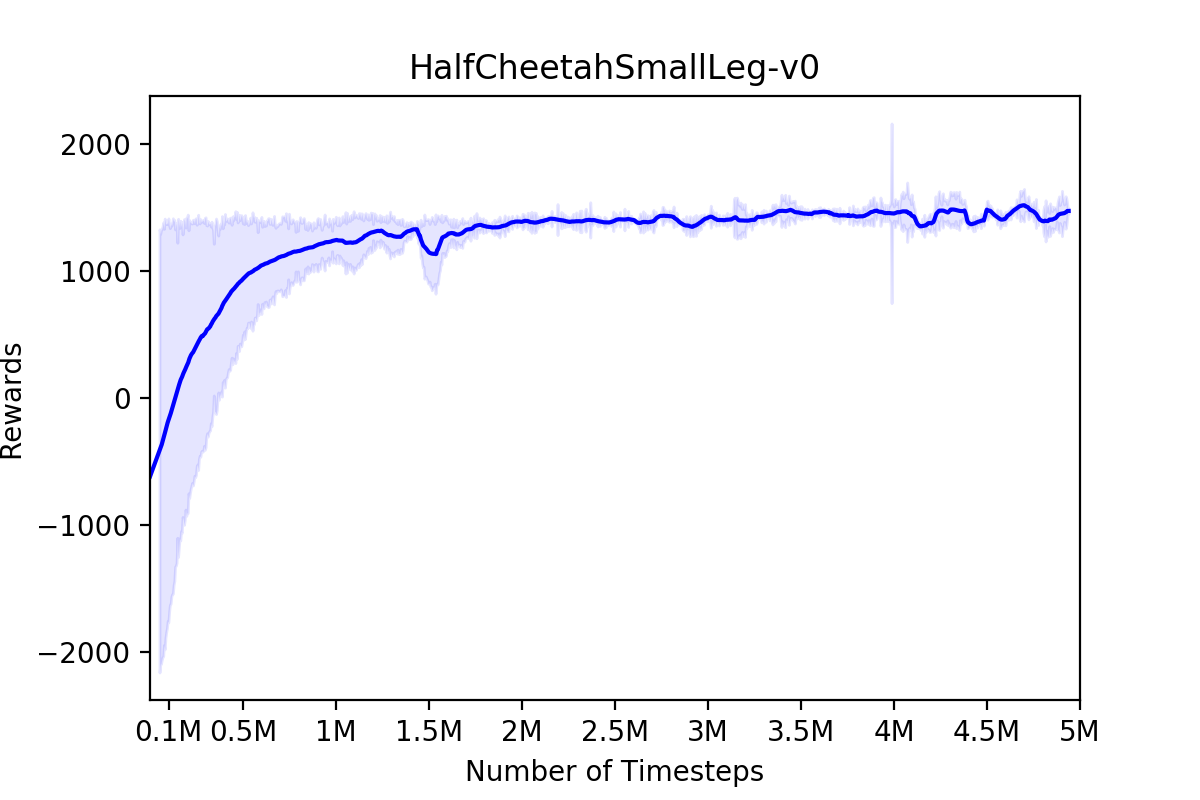}
\caption{}
\end{subfigure}
\hfil
\begin{subfigure}[t]{0.48\linewidth}
\centering
\includegraphics[width=0.8\linewidth]{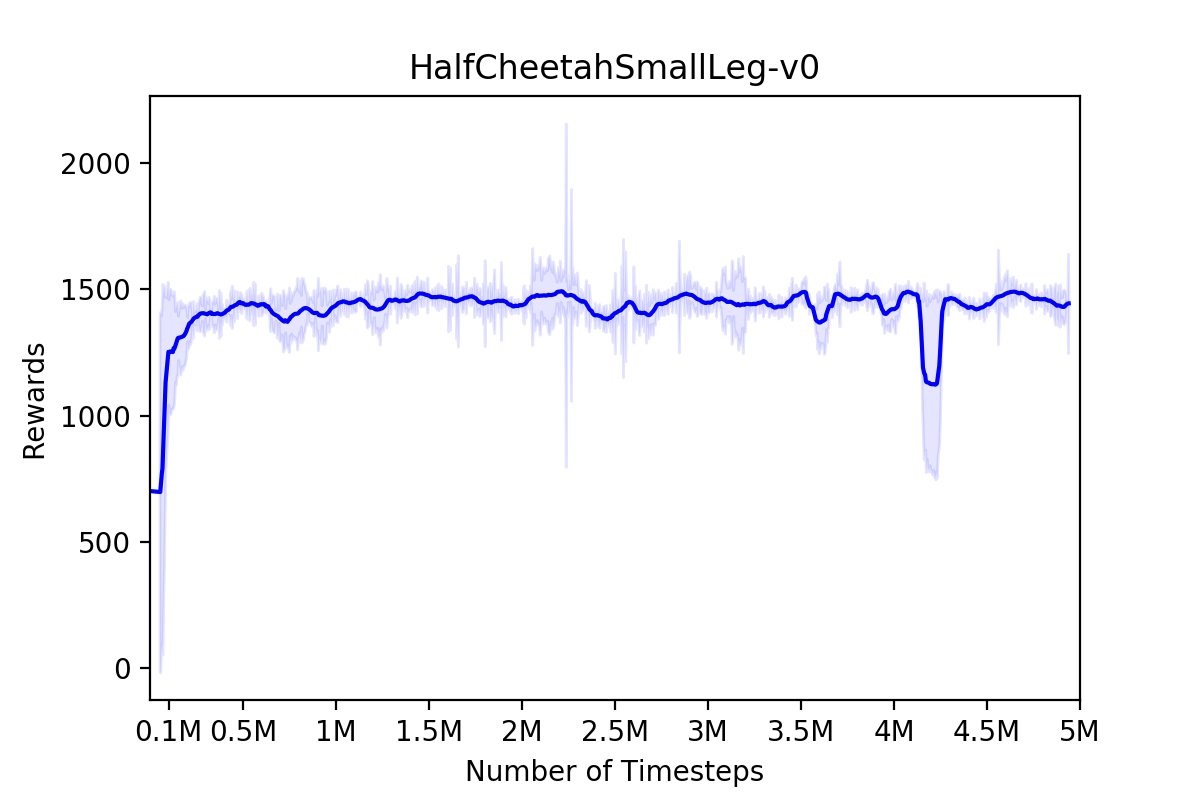}
\caption{}
\end{subfigure}
\\
\begin{subfigure}[t]{0.48\linewidth}
\centering
\includegraphics[width=0.8\linewidth]{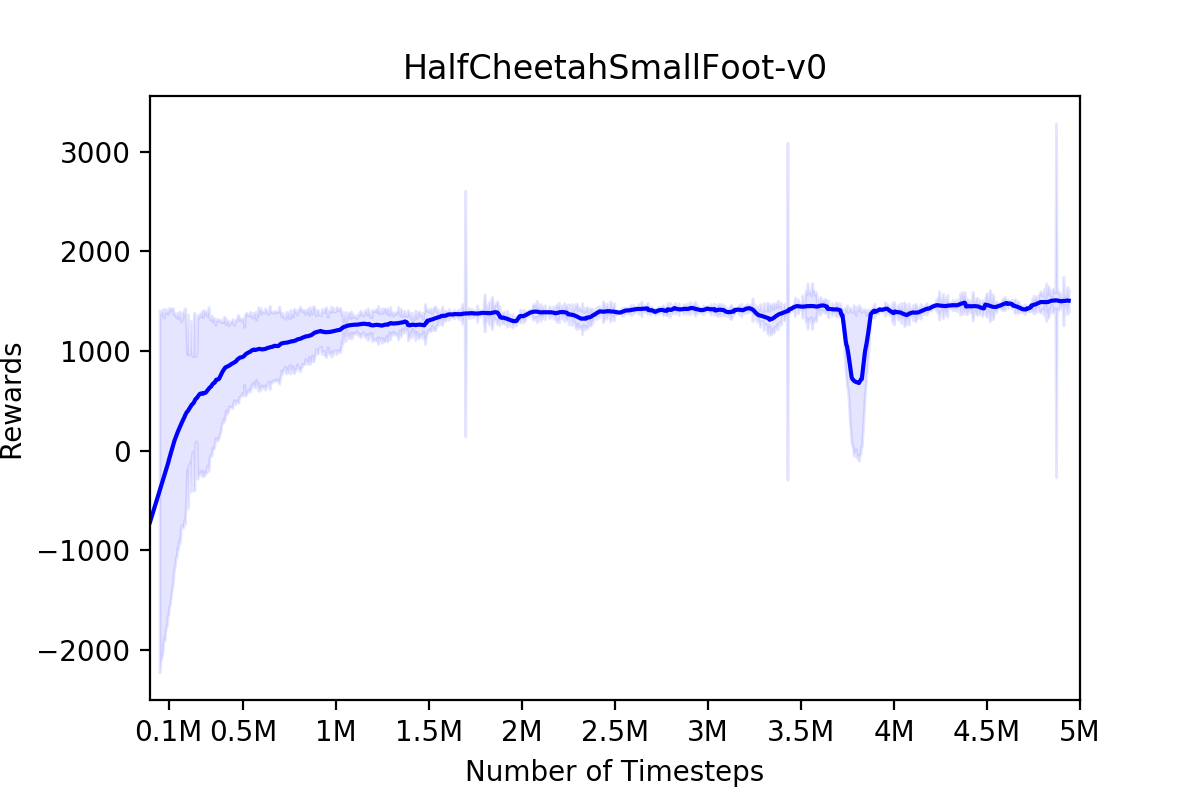}
\caption{}
\end{subfigure}
\hfil
\begin{subfigure}[t]{0.48\linewidth}
\centering
\includegraphics[width=0.8\linewidth]{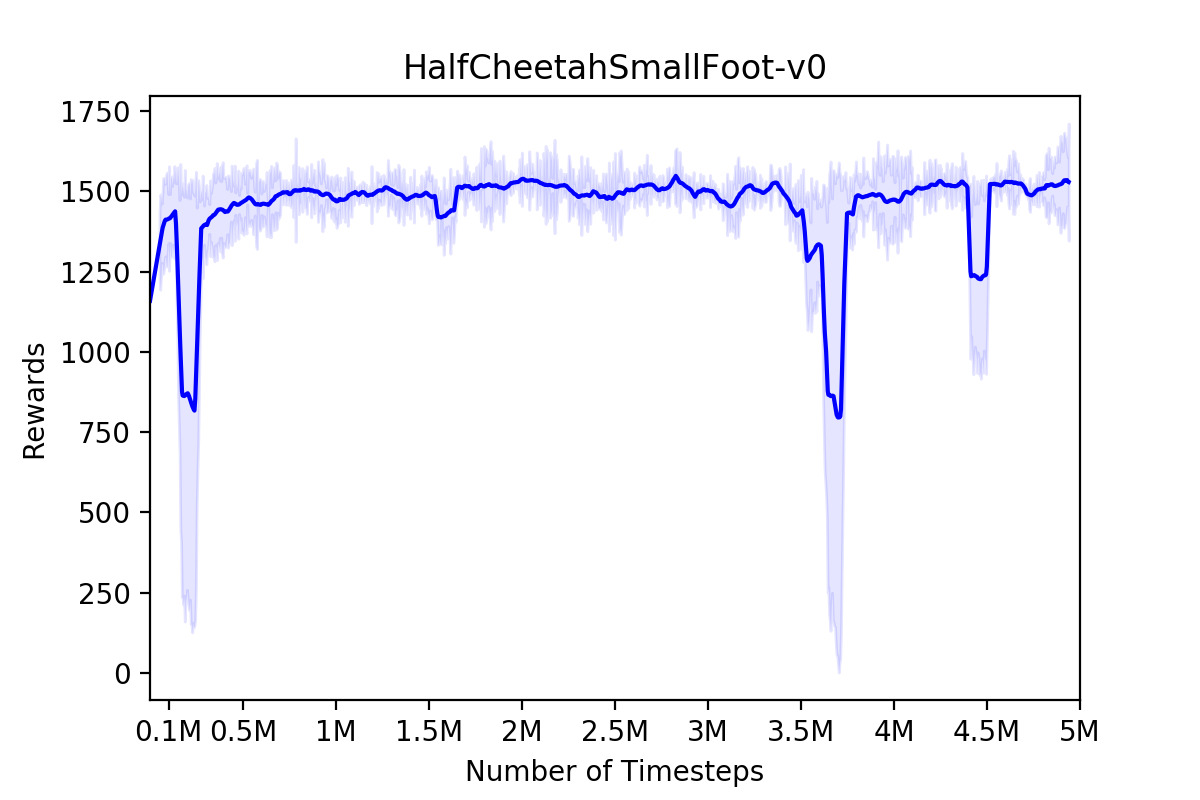}
\caption{}
\end{subfigure}
\caption{Rewards computed across timesteps during training. $(a)$ environment \textit{HalfCheetahSmallLeg-v0} trained from scratch, $(b)$ environment \textit{HalfCheetahSmallLeg-v0} fine tuned with transferred weights from \textit{HalfCheetahSmallFoot-v0}, $(c)$ \textit{HalfCheetahSmallFoot-v0} environment trained from scratch, $(d)$ environment \textit{HalfCheetahSmallFoot-v0} fine tuned with transferred weights from \textit{HalfCheetahSmallLeg-v0 }}
\label{finetune_fig}
\end{figure}

\section{Multi-Task graphs}


\begin{figure}[h!]
\centering
\begin{subfigure}[t]{0.32\linewidth}
\centering
\includegraphics[width=\linewidth]{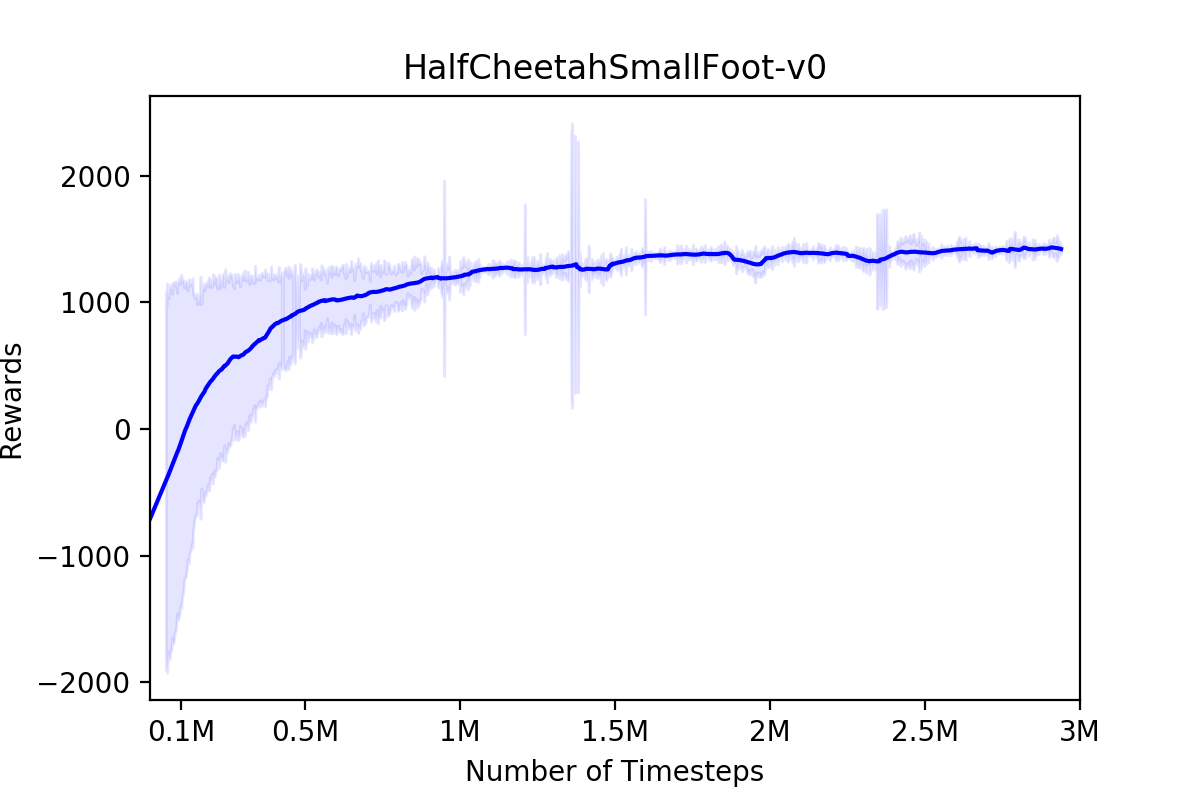}
\caption{\scriptsize{Single-task training from scratch}}
\end{subfigure}
\hfil
\begin{subfigure}[t]{0.32\linewidth}
\centering
\includegraphics[width=\linewidth]{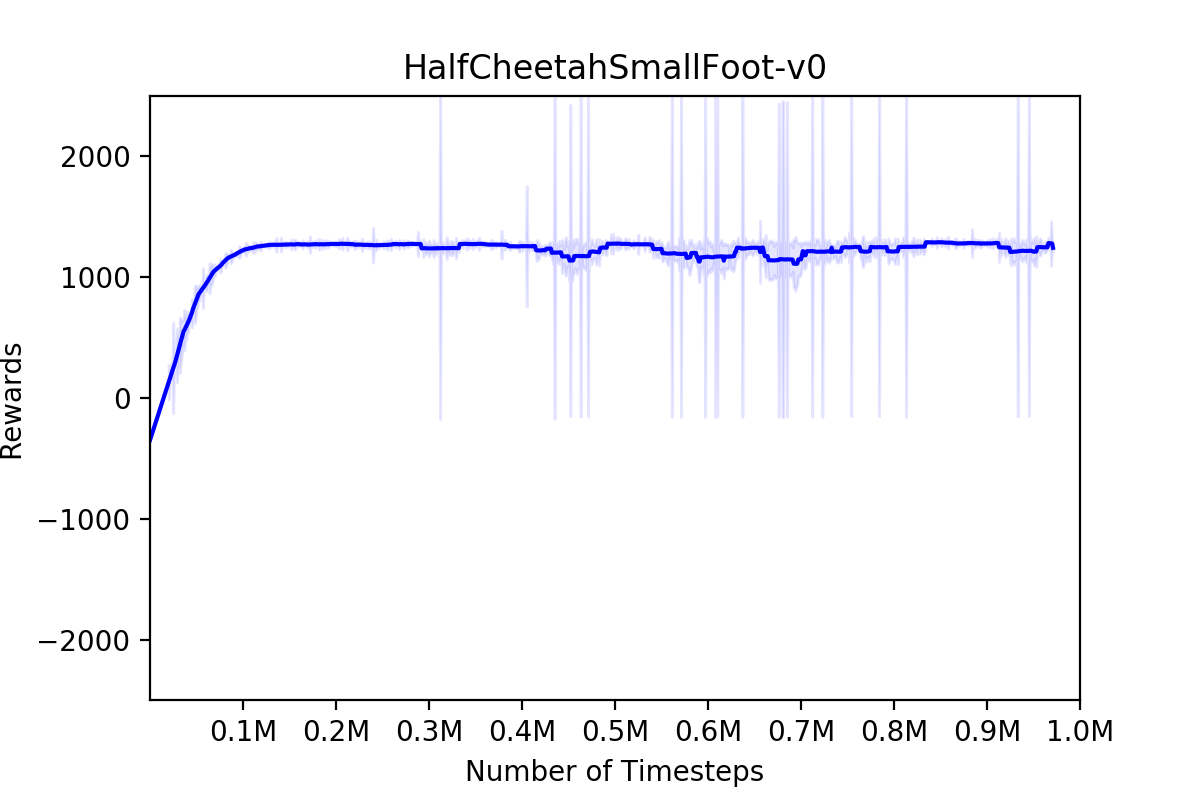}
\caption{\scriptsize{6-task training using distillation}}
\end{subfigure}
\hfil
\begin{subfigure}[t]{0.32\linewidth}
\centering
\includegraphics[width=\linewidth]{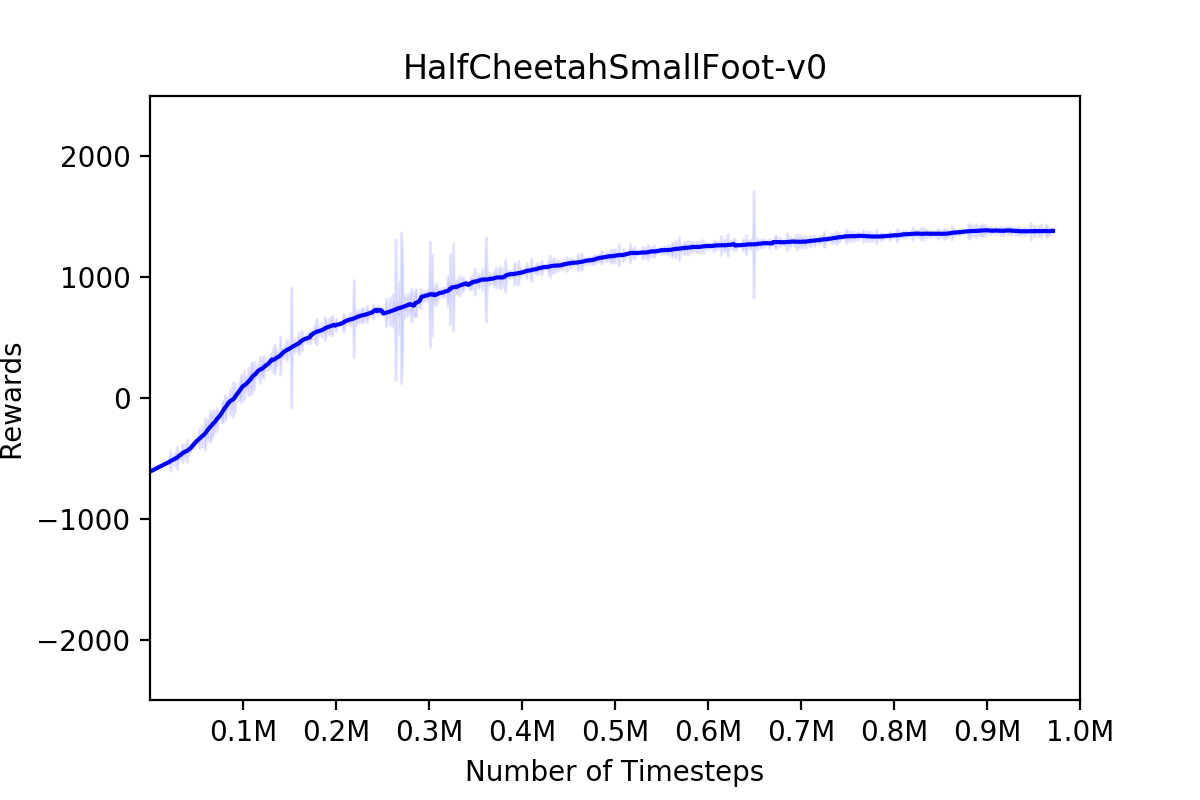}
\caption{\scriptsize{6-task training from scratch}}
\end{subfigure}
\caption{Learning curves for \textit{HalfCheetahSmallFoot-v0}}
\end{figure}


\begin{figure}[h!]
\centering
\begin{subfigure}[t]{0.32\linewidth}
\centering
\includegraphics[width=\linewidth]{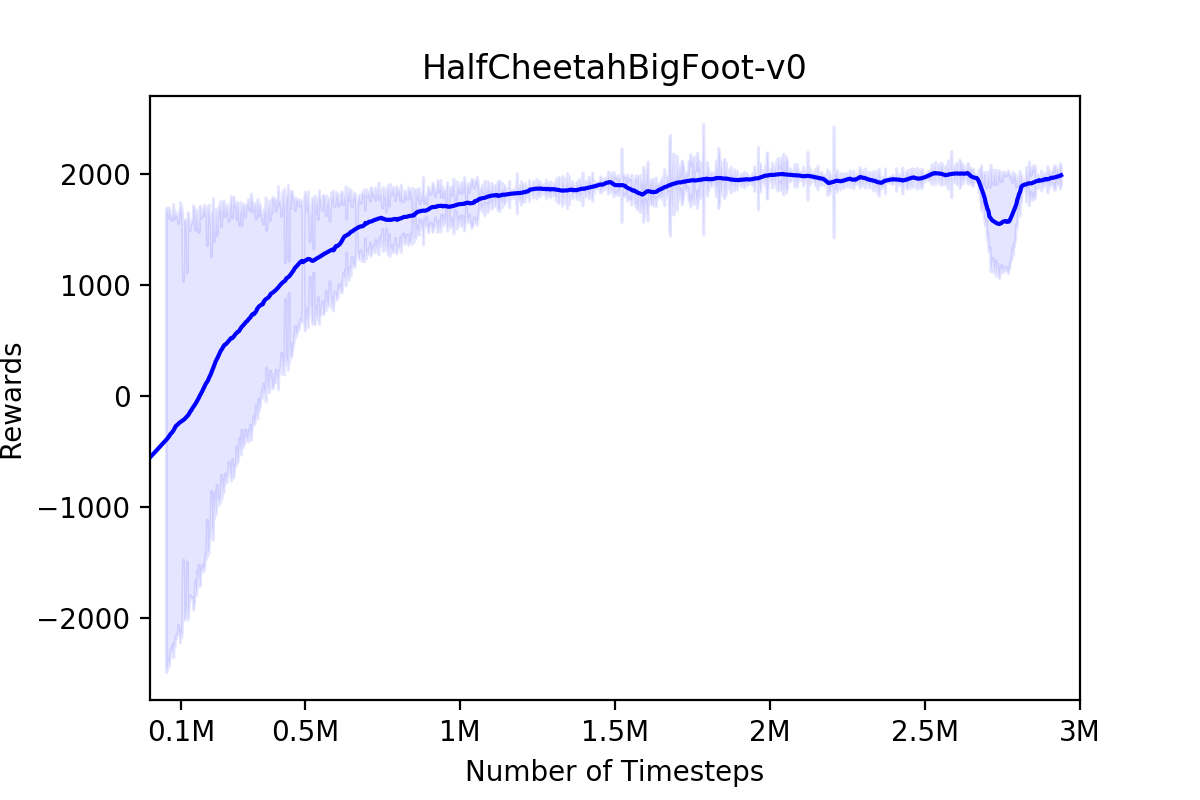}
\caption{\scriptsize{Single-task training from scratch}}
\end{subfigure}
\hfil
\begin{subfigure}[t]{0.32\linewidth}
\centering
\includegraphics[width=\linewidth]{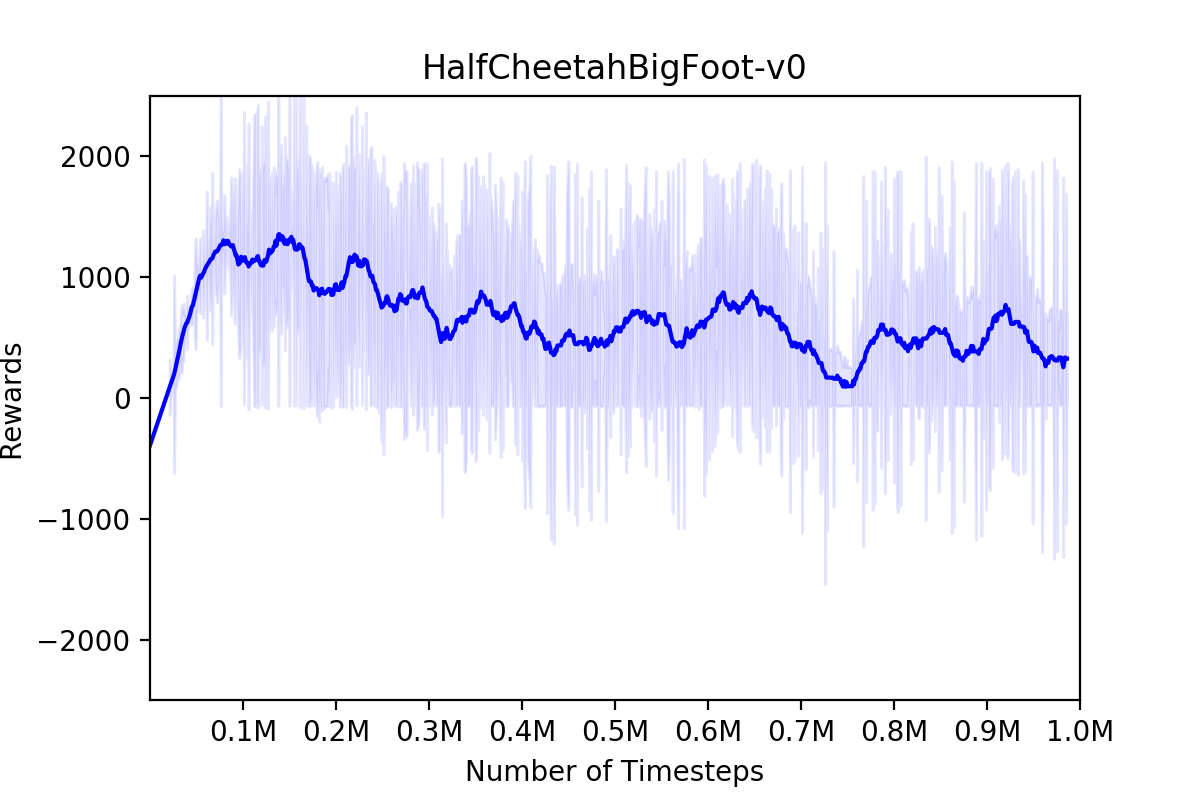}
\caption{\scriptsize{6-task training using distillation}}
\end{subfigure}
\hfil
\begin{subfigure}[t]{0.32\linewidth}
\centering
\includegraphics[width=\linewidth]{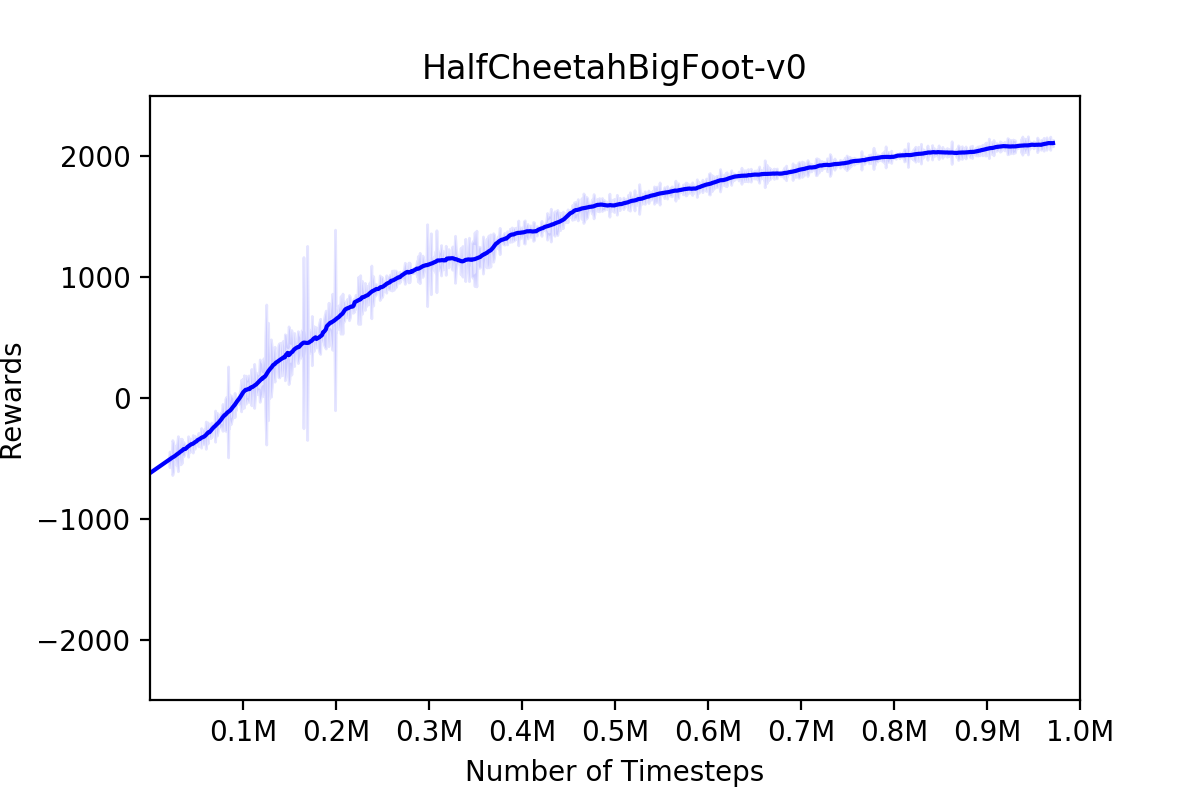}
\caption{\scriptsize{6-task training from scratch}}
\end{subfigure}
\caption{Learning curves for \textit{HalfCheetahBigFoot-v0}}
\end{figure}


\begin{figure}[h!]
\centering
\begin{subfigure}[t]{0.32\linewidth}
\centering
\includegraphics[width=\linewidth]{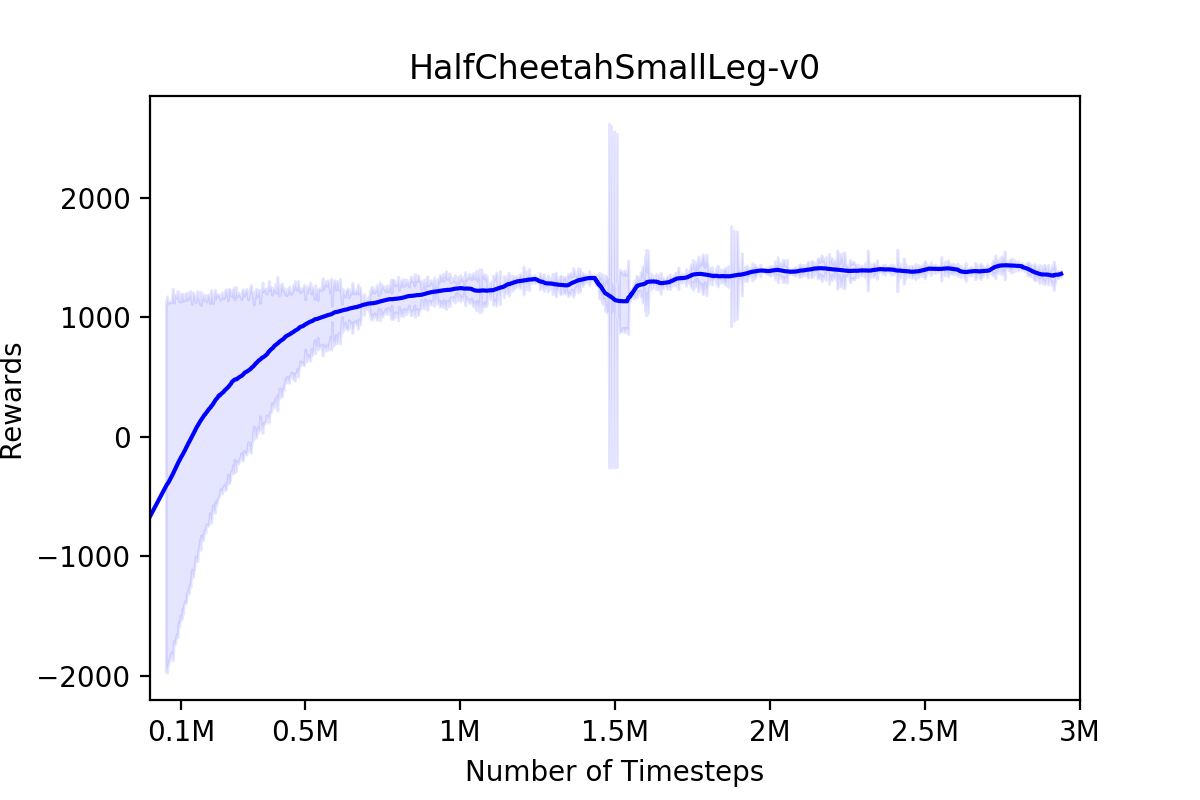}
\caption{\scriptsize{Single-task training from scratch}}
\end{subfigure}
\hfil
\begin{subfigure}[t]{0.32\linewidth}
\centering
\includegraphics[width=\linewidth]{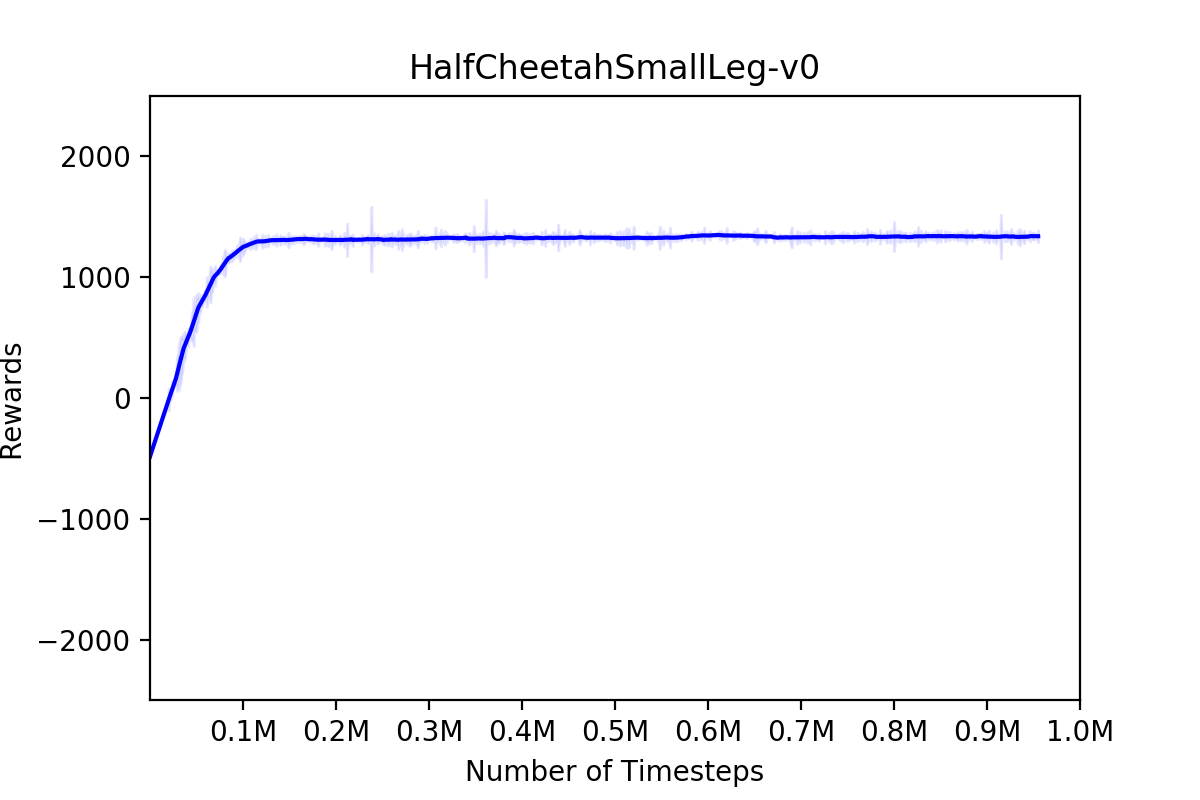}
\caption{\scriptsize{6-task training using distillation}}
\end{subfigure}
\hfil
\begin{subfigure}[t]{0.32\linewidth}
\centering
\includegraphics[width=\linewidth]{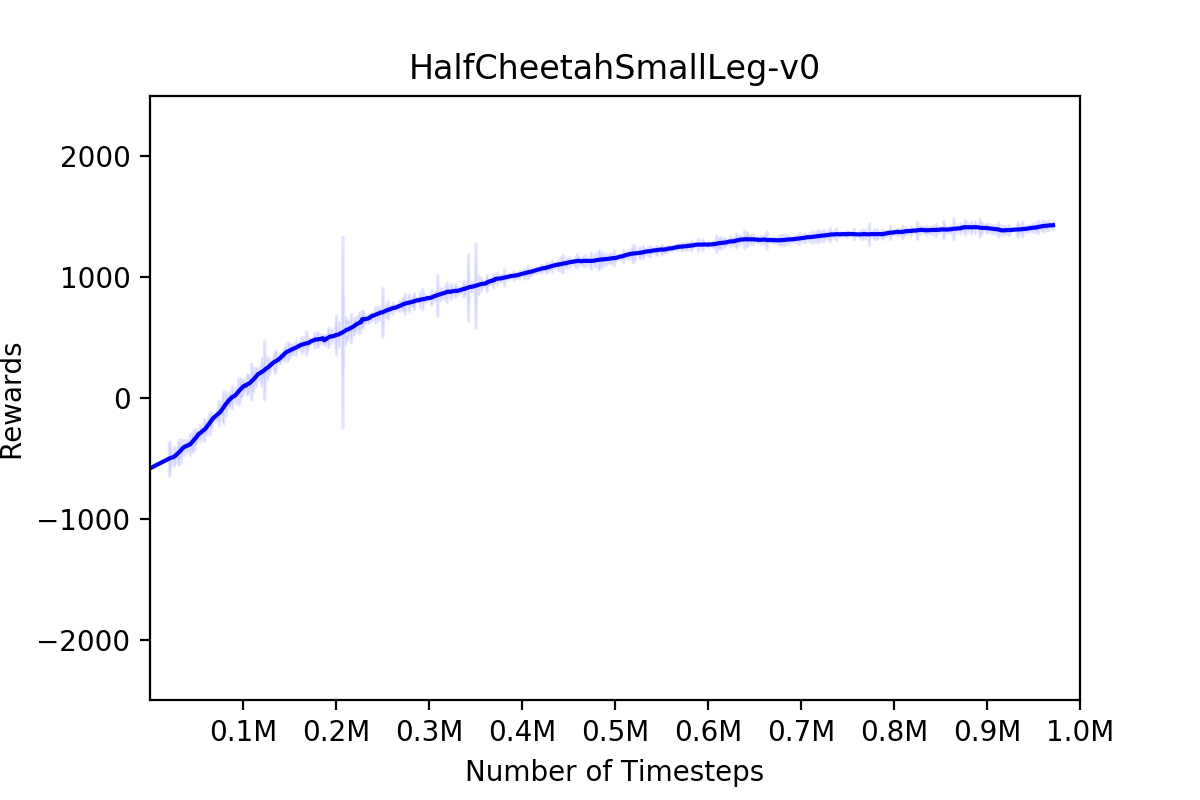}
\caption{\scriptsize{6-task training from scratch}}
\end{subfigure}
\caption{Learning curves for \textit{HalfCheetahSmallLeg-v0}}
\end{figure}


\begin{figure}[h!]
\centering
\begin{subfigure}[t]{0.32\linewidth}
\centering
\includegraphics[width=\linewidth]{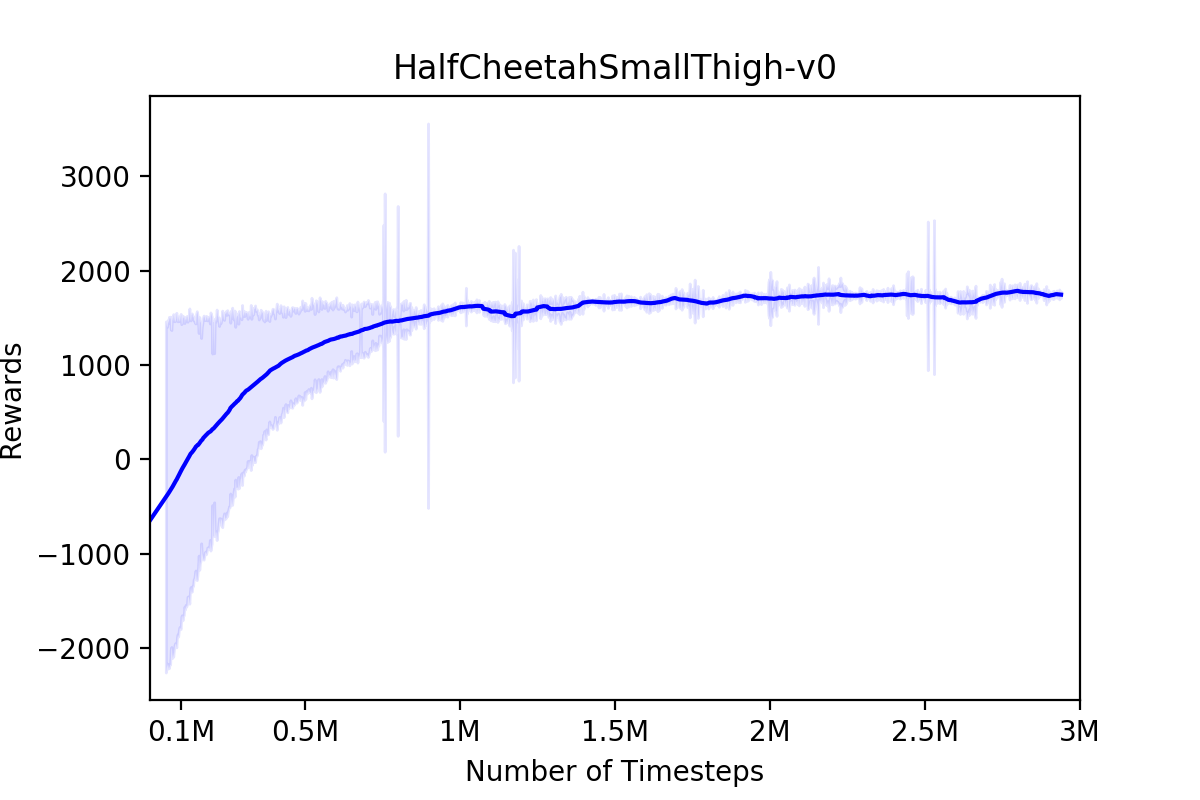}
\caption{\scriptsize{Single-task training from scratch}}
\end{subfigure}
\hfil
\begin{subfigure}[t]{0.32\linewidth}
\centering
\includegraphics[width=\linewidth]{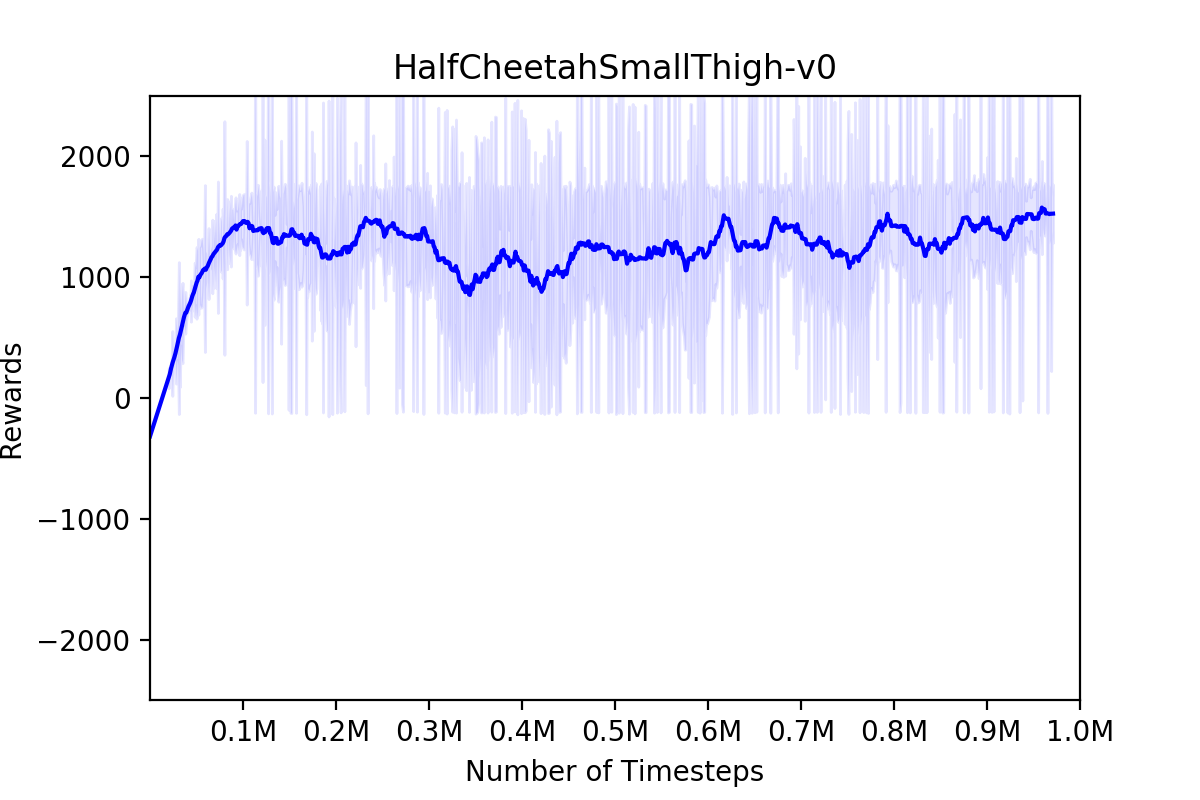}
\caption{\scriptsize{6-task training using distillation}}
\end{subfigure}
\hfil
\begin{subfigure}[t]{0.32\linewidth}
\centering
\includegraphics[width=\linewidth]{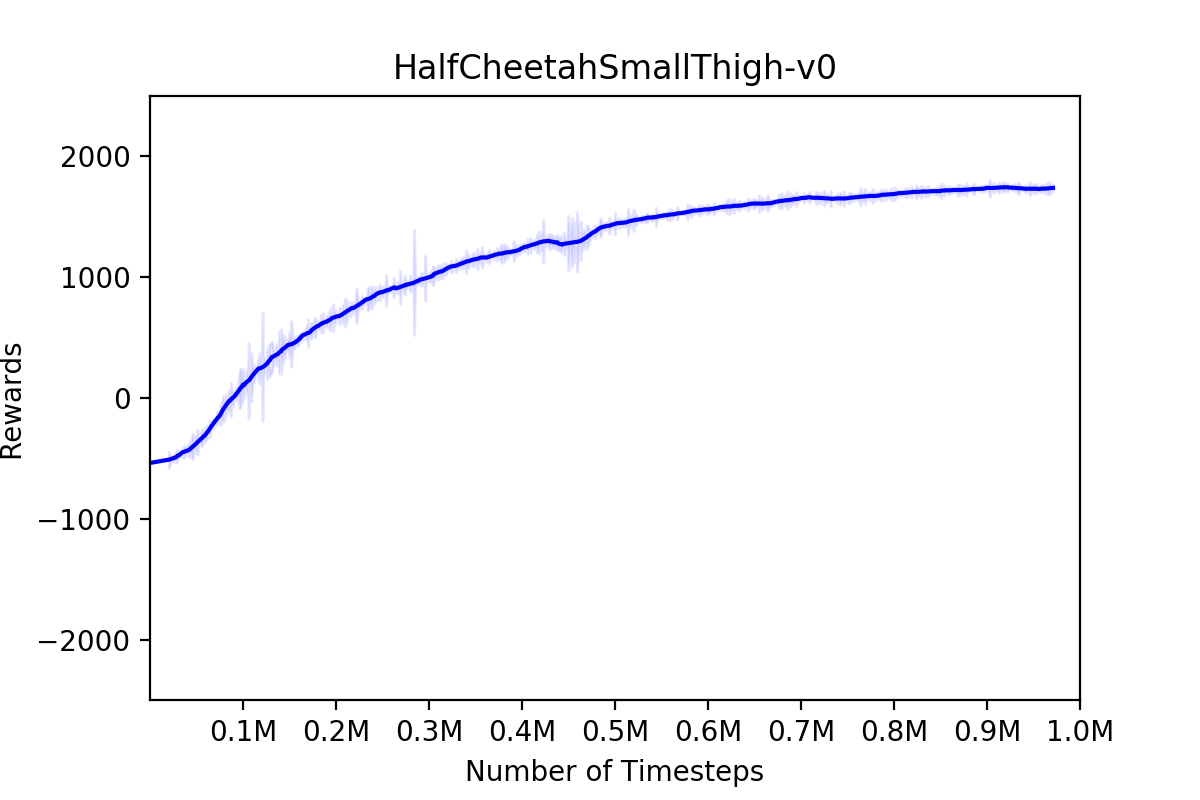}
\caption{\scriptsize{6-task training from scratch}}
\end{subfigure}
\caption{Learning curves for \textit{HalfCheetahSmallThigh-v0}}
\end{figure}


\begin{figure}[h!]
\centering
\begin{subfigure}[t]{0.32\linewidth}
\centering
\includegraphics[width=\linewidth]{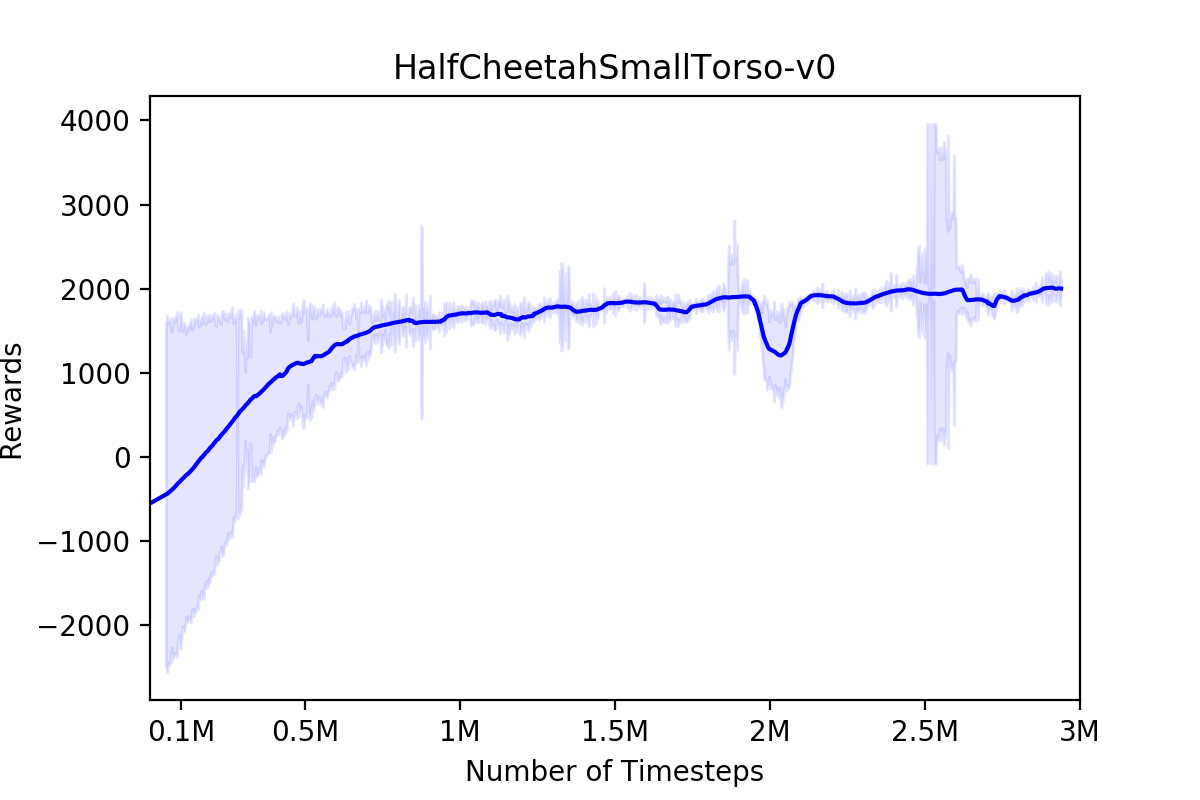}
\caption{\scriptsize{Single-task training from scratch}}
\end{subfigure}
\hfil
\begin{subfigure}[t]{0.32\linewidth}
\centering
\includegraphics[width=\linewidth]{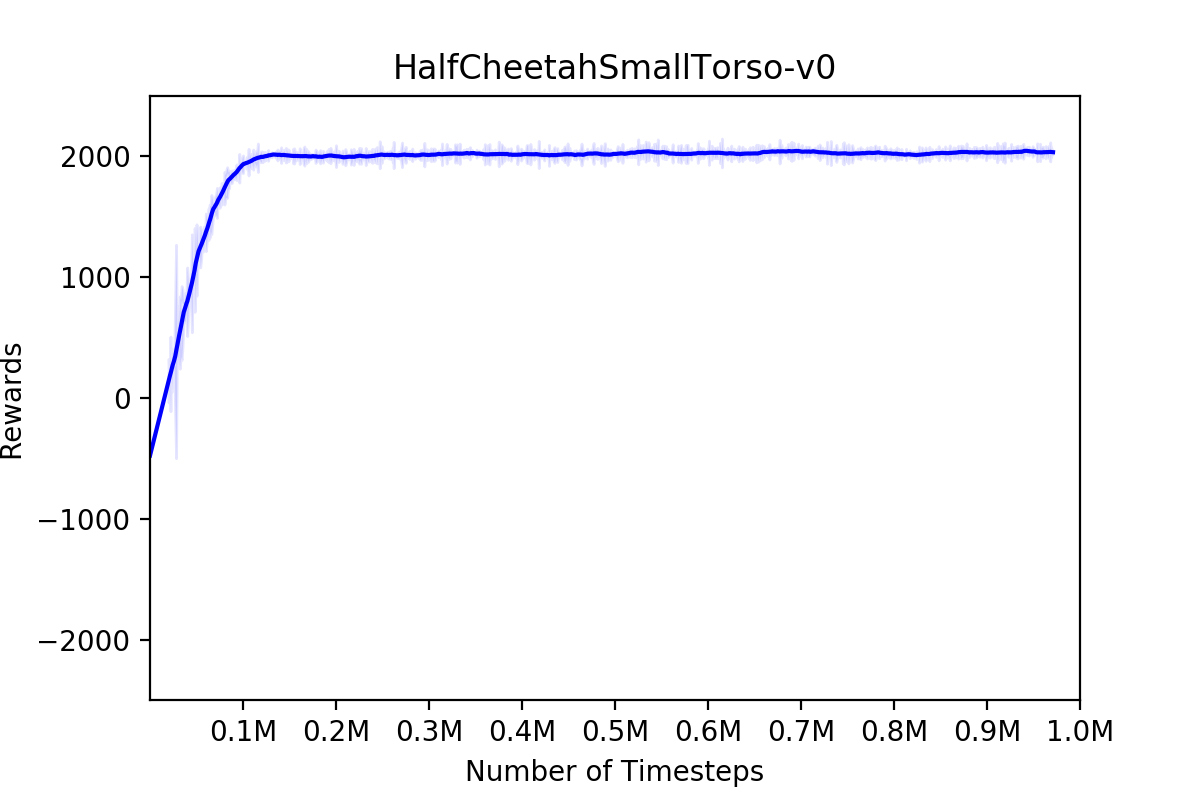}
\caption{\scriptsize{6-task training using distillation}}
\end{subfigure}
\hfil
\begin{subfigure}[t]{0.32\linewidth}
\centering
\includegraphics[width=\linewidth]{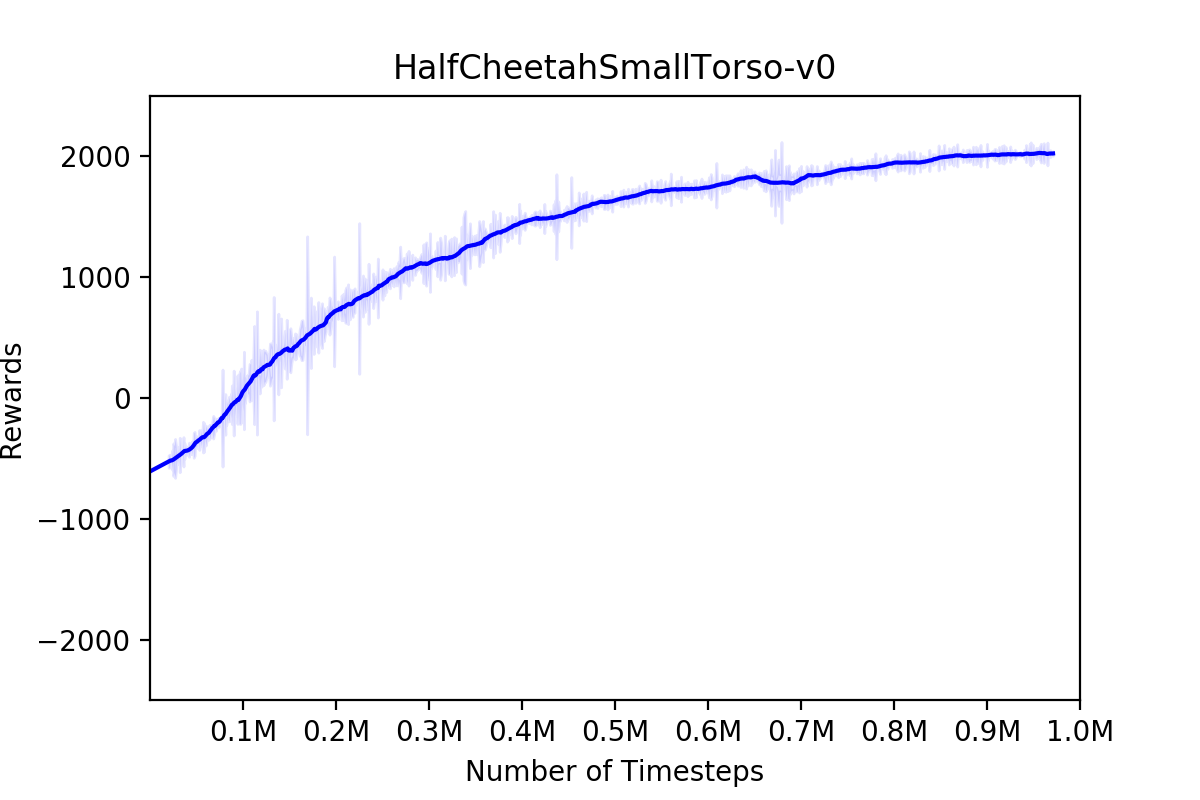}
\caption{\scriptsize{6-task training from scratch}}
\end{subfigure}
\caption{Learning curves for \textit{HalfCheetahSmallTorso-v0}}
\end{figure}


\begin{figure}[h!]
\centering
\begin{subfigure}[t]{0.32\linewidth}
\centering
\includegraphics[width=\linewidth]{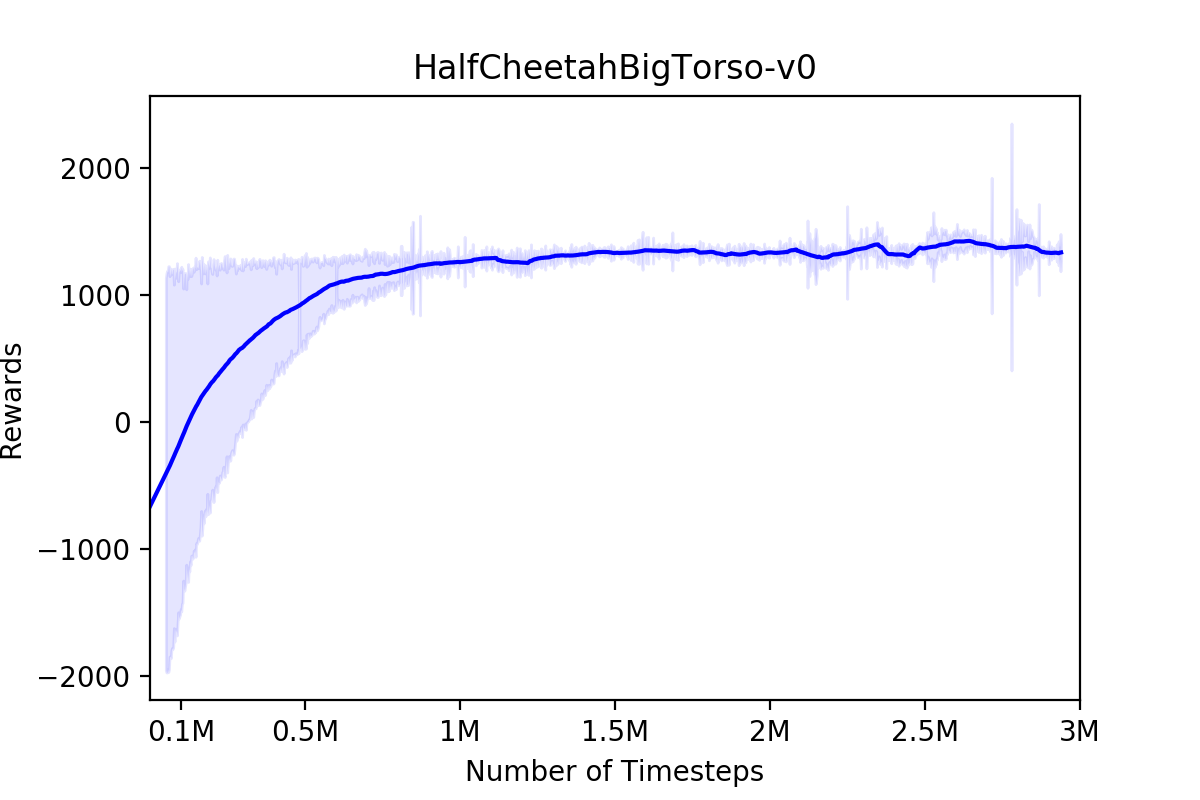}
\caption{\scriptsize{Single-task training from scratch}}
\end{subfigure}
\hfil
\begin{subfigure}[t]{0.32\linewidth}
\centering
\includegraphics[width=\linewidth]{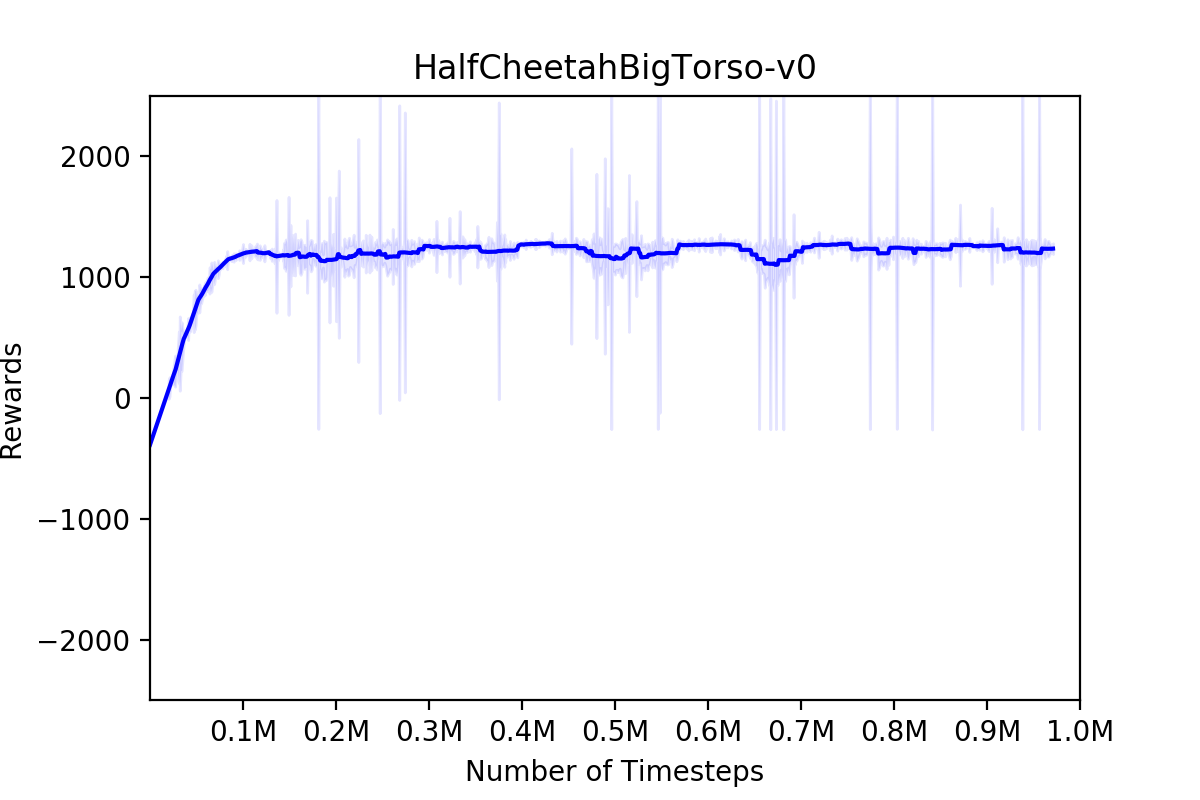}
\caption{\scriptsize{6-task training using distillation}}
\end{subfigure}
\hfil
\begin{subfigure}[t]{0.32\linewidth}
\centering
\includegraphics[width=\linewidth]{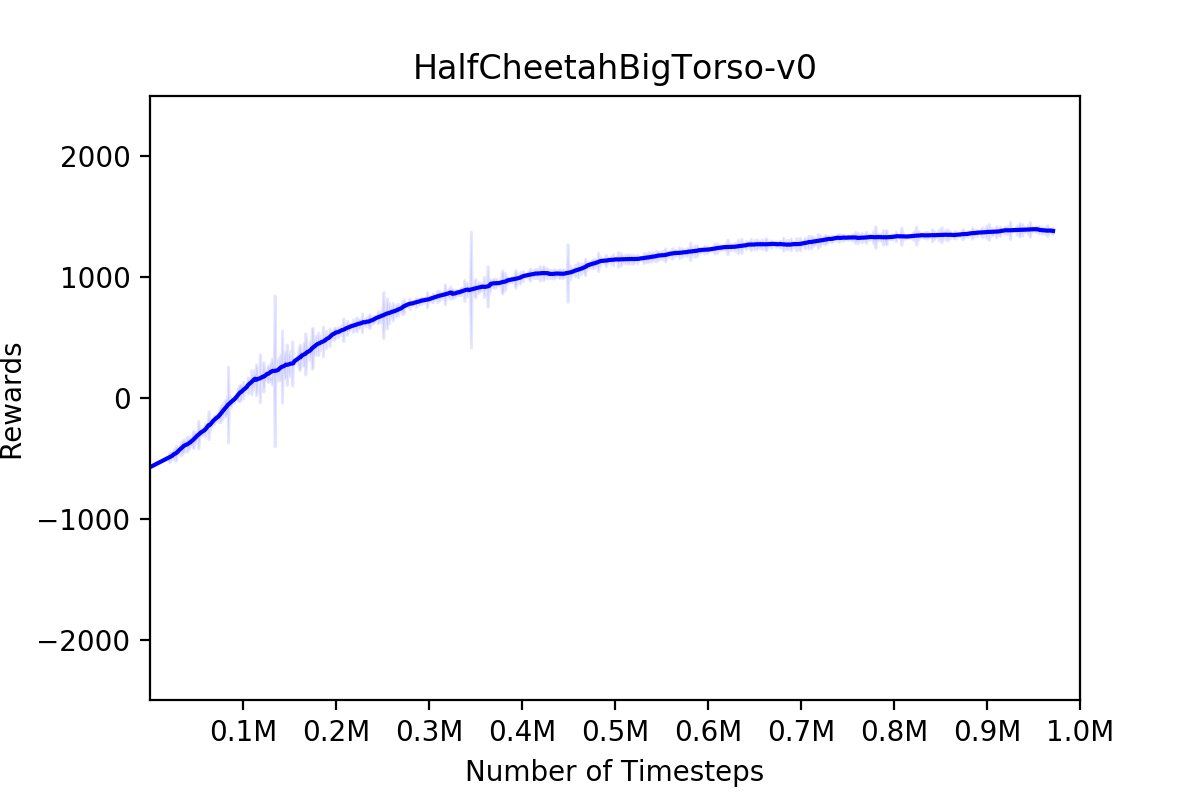}
\caption{\scriptsize{6-task training from scratch}}
\end{subfigure}
\caption{Learning curves for \textit{HalfCheetahBigTorso-v0}}
\end{figure}

\twocolumn

\end{document}